\documentclass[pdflatex,sn-mathphys-num]{sn-jnl}

\usepackage{graphicx}%
\usepackage{multirow}%
\usepackage{amsmath,amssymb,amsfonts}%
\usepackage{amsthm}%
\usepackage{mathrsfs}%
\usepackage[title]{appendix}%
\usepackage{xcolor}%
\usepackage{textcomp}%
\usepackage{manyfoot}%
\usepackage{booktabs}%
\usepackage{algorithm}%
\usepackage{algorithmicx}%
\usepackage{algpseudocode}%
\usepackage{listings}%
\usepackage[T1]{fontenc}
\usepackage[utf8]{inputenc}

\usepackage{tcolorbox}
\usepackage{enumitem}

\newtcolorbox{examplebox}{
  colback=gray!5,
  colframe=gray!60,
  boxrule=0.4pt,
  arc=2pt,
  left=6pt,right=6pt,top=6pt,bottom=6pt
}

\theoremstyle{thmstyleone}%
\theoremstyle{thmstyletwo}%

\theoremstyle{thmstylethree}%

\begin{document}

\title[Epistemic orientation in parliamentary discourse is associated with deliberative democracy]{Epistemic orientation in parliamentary discourse is associated with deliberative democracy}

\author*[1]{\fnm{Segun} \sur{Aroyehun}}\email{segun.aroyehun@uni-konstanz.de}
\author[2,4]{\fnm{Stephan} \sur{Lewandowsky}} 
\author[1,3]{\fnm{David} \sur{Garcia}} 

\affil*[1]{\orgname{Department of Politics and Public Administration, University of Konstanz}, \country{Germany}}

\affil[2]{\orgname{School of Psychological Science, University of Bristol}, \country{UK}}

\affil[3]{\orgname{Complexity Science Hub}, \country{Austria}}

\affil[4]{\orgname{Department of Psychology, University of Potsdam}, \country{Germany}}

\abstract{The pursuit of truth is central to democratic deliberation and governance, yet political discourse reflects varying epistemic orientations, ranging from evidence-based reasoning grounded in verifiable information to intuition-based reasoning rooted in beliefs and subjective interpretation. We introduce a scalable approach to measure epistemic orientation using the Evidence–Minus–Intuition (EMI) score, derived from large language model (LLM) ratings and embedding-based semantic similarity. Applying this approach to 15 million parliamentary speech segments spanning 1946 to 2025 across seven countries, we examine temporal patterns in discourse and its association with deliberative democracy and governance. We find that EMI is positively associated with deliberative democracy within countries over time, with consistent relationships in both contemporaneous and lagged analyses. EMI is also positively associated with the transparency and predictable implementation of laws as a dimension of governance. These findings suggest that the epistemic nature of political discourse is crucial for both the quality of democracy and governance.}



\maketitle
\section*{Main text}\label{sec1}

Truth underpins accountability, transparency, and informed decision-making in democratic societies, providing a shared basis for evaluating competing claims in parliamentary debate. Truth, however, can be pursued through evidence-based reasoning grounded in facts and data, or, people can engage in intuition-based reasoning grounded in beliefs, values, and subjective interpretation \cite{lewandowsky2020willful, cooper2023honest}. Productive democratic discourse requires a balance between these approaches. Evidence enables the adjudication of competing positions, while intuition contributes moral and experiential dimensions that are often indispensable in public life.

Recent computational analyses of parliamentary discourse in the United States document a decline in evidence-based language in congressional speeches relative to intuition-based language over time \cite{Aroyehun2024Computational}, raising concerns about the erosion of epistemic norms that underpin democratic deliberation. This development also raises a broader question: How does parliamentary language relate to the health of a country's democracy? To address this question, prior work has introduced an Evidence-Minus-Intuition (EMI) score as a measure of the epistemic orientation of discourse that can be applied at scale. The epistemic orientation captured by EMI is relevant to how political systems deliberate and act. 
Related research extends the analysis beyond parliamentary settings to elite communication \cite{lasser2023alternative} and elite–citizen interaction \cite{carrella_different_2025} on social media. Across these contexts, more evidence-oriented discourse is associated with the sharing of higher-quality information \cite{lasser2023alternative}, less polarizing subsequent discussion \cite{carrella_different_2025}, and greater legislative productivity \cite{Aroyehun2024Computational}.

However, existing research remains limited to one institutional and linguistic context. Therefore, it is unclear whether EMI reflects the nature of political discourse and the broader dimensions of democratic and governance quality across countries, political systems, and languages. A key reason for this limitation has been methodological. Earlier approaches relied on language-specific dictionaries or computational resources, constraining comparative analysis. Recent advances in natural language processing, particularly the development of LLMs, enable the analysis of political discourse across languages and contexts in a more comparable and scalable manner. This study builds on these advances by extending the analysis of EMI beyond the United States. We develop an approach to measure the epistemic orientation of parliamentary speeches across multiple countries, allowing for the systematic study of discourse across institutional and linguistic settings. Our focus is on how political rhetoric expresses the pursuit of truth, rather than on evaluating the truth value of individual statements.

We apply computational text analysis based on LLMs \cite{rathje2024gpt, Aroyehun2024Computational, jha2025does} to measure the relative salience of evidence-based and intuition-based language in text segments from parliamentary transcripts spanning 1946 to 2025 across seven countries: United States of America, West Germany, Germany, Italy, Iceland, Poland, and Turkey. These countries cover a range of different political systems, from liberal democracies, electoral democracies, to a competitive autocracy.
We use the Deliberative Democracy Index (DDI) \cite{Coppedge2026Vdem}, a composite measure based on expert assessments of electoral and deliberative aspects of democracy including reasoned justification, appeals to the common good, and respect for counterarguments, to assess the health of a country's democracy. We examine the association between parliamentary EMI and the DDI. Rhetoric may also shape the quality of governance \cite{grant2008legislative, Aroyehun2024Computational}, we additionally examine the relationship between EMI and the V-Dem index of transparent laws and predictable enforcement (TPL) \cite{Coppedge2026Vdem}, which captures whether laws are transparent, coherent, and predictably enforced. 

\subsection*{LLM-based measurement of evidence-based and intuition-based language}

Our analysis is based on 15 million text segments drawn from parliamentary transcripts spanning 1946 to 2025 across seven countries (see Methods for details).
We measure the relative salience of evidence-based versus intuition-based language using the EMI score \cite{Aroyehun2024Computational}. We combine two complementary approaches: (i) ratings from three large language models (LLMs) \cite{rathje2024gpt, plaza2024wisdom} and (ii) semantic similarity measures derived from contextual language embeddings, which represent text segments and conceptual anchors in a shared vector space \cite{Aroyehun2024Computational,jha2025does}. 

For the embedding-based component, we construct semantic anchors for evidence-based and intuition-based language using 15 seed terms per category, derived from prior work \cite{Aroyehun2024Computational} and expanded with dictionary definitions. To enable multilingual analysis, anchors are translated into each target language and reviewed for consistency.

EMI is computed as the difference between evidence- and intuition-based scores in each approach: LLM ratings (average difference) and embedding similarity (difference in cosine similarity to anchors). 
Positive values indicate greater use of evidence-based language, negative values indicate greater reliance on intuition-based language. Examples and validation against human annotations are provided in the Methods.

\subsection*{Temporal patterns of EMI and DDI}

\begin{figure}[htbp]
    \centering    \includegraphics[width=0.95\linewidth, height=0.85\textheight]{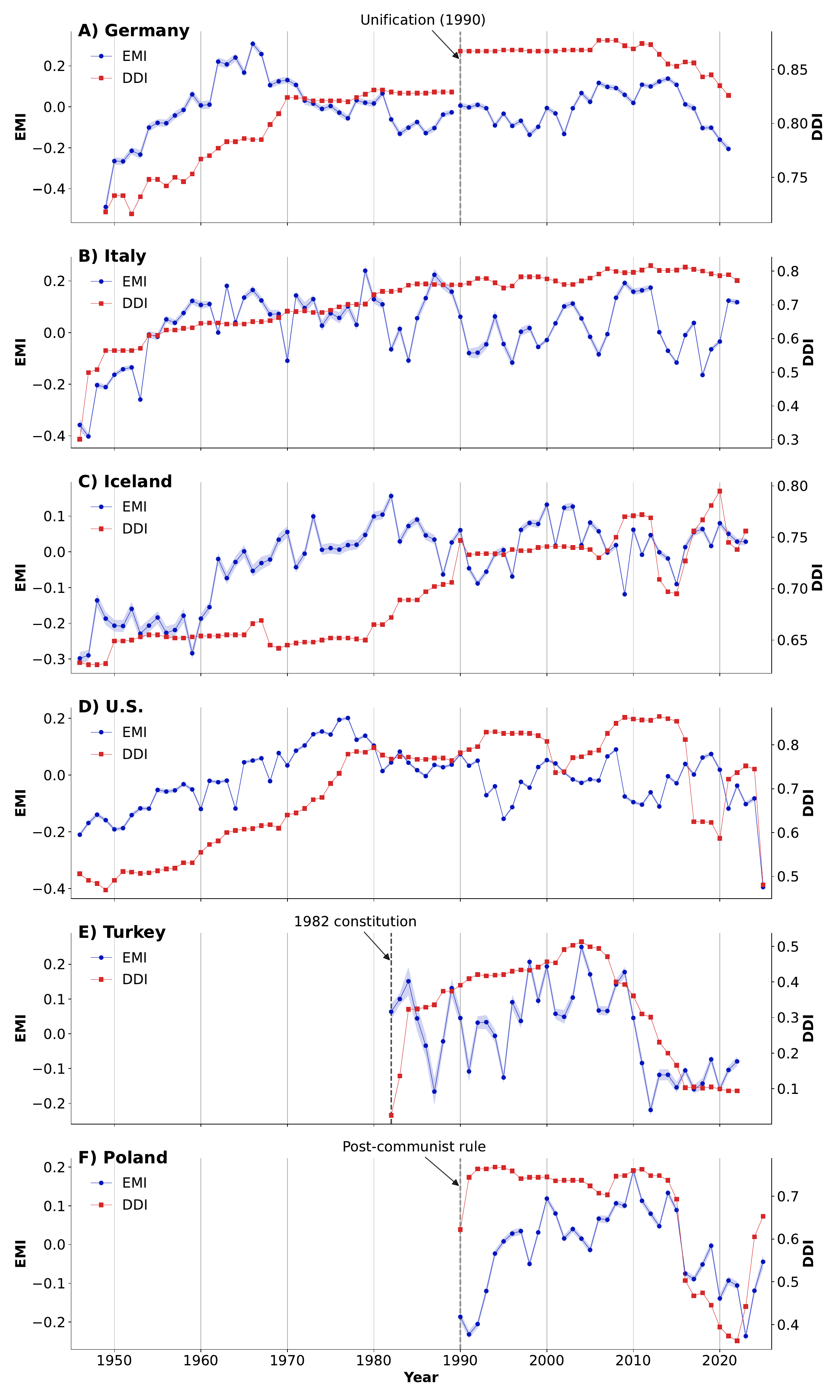}
    \caption{Plots showing trends in yearly observations of EMI and DDI across countries. Panels show (A) Germany (West Germany pre-1990 and unified Germany thereafter), (B) Italy, (C) Iceland, (D) United States, (E) Turkey, and (F) Poland. EMI is shown with 95\% confidence interval bands based on 10,000 bootstrap iterations. Intervals may be too small to be visible due to the large sample size. Vertical dashed lines mark major institutional changes: in Germany, unification in 1990; in Turkey, the adoption of the 1982 constitution; and in Poland, the post-communist transition.}
    \label{fig:emi_ddi_trend}
\end{figure}

Figure \ref{fig:emi_ddi_trend} shows trajectories of epistemic orientation (EMI) and deliberative democracy (DDI) across countries over time. 
In West Germany, DDI rises steadily postwar and stabilizes at a high level, reflecting institutional consolidation. EMI increases early and then fluctuates within a narrow range. After reunification, DDI exhibits a step increase likely linked to full sovereignty, motivating separate analysis of pre- and post-1989 Germany. Post-unification, DDI remains high, while EMI becomes more variable, potentially reflecting the emergence of new challenger parties.

In Italy, DDI increases gradually, while EMI remains highly volatile, consistent with government turnover and shifts in party competition since the 1990s. In the United States, DDI rises postwar and remains high; EMI increases until the late twentieth century, then becomes more volatile with a recent decline, consistent with polarization trends \cite{Aroyehun2024Computational}. Iceland shows steady increases in both measures with low volatility. In Poland and Turkey, both EMI and DDI rise following regime transitions and decline in recent years, coinciding with broader political changes.

\begin{figure}[htbp]
    \centering
    \includegraphics[scale=0.4]{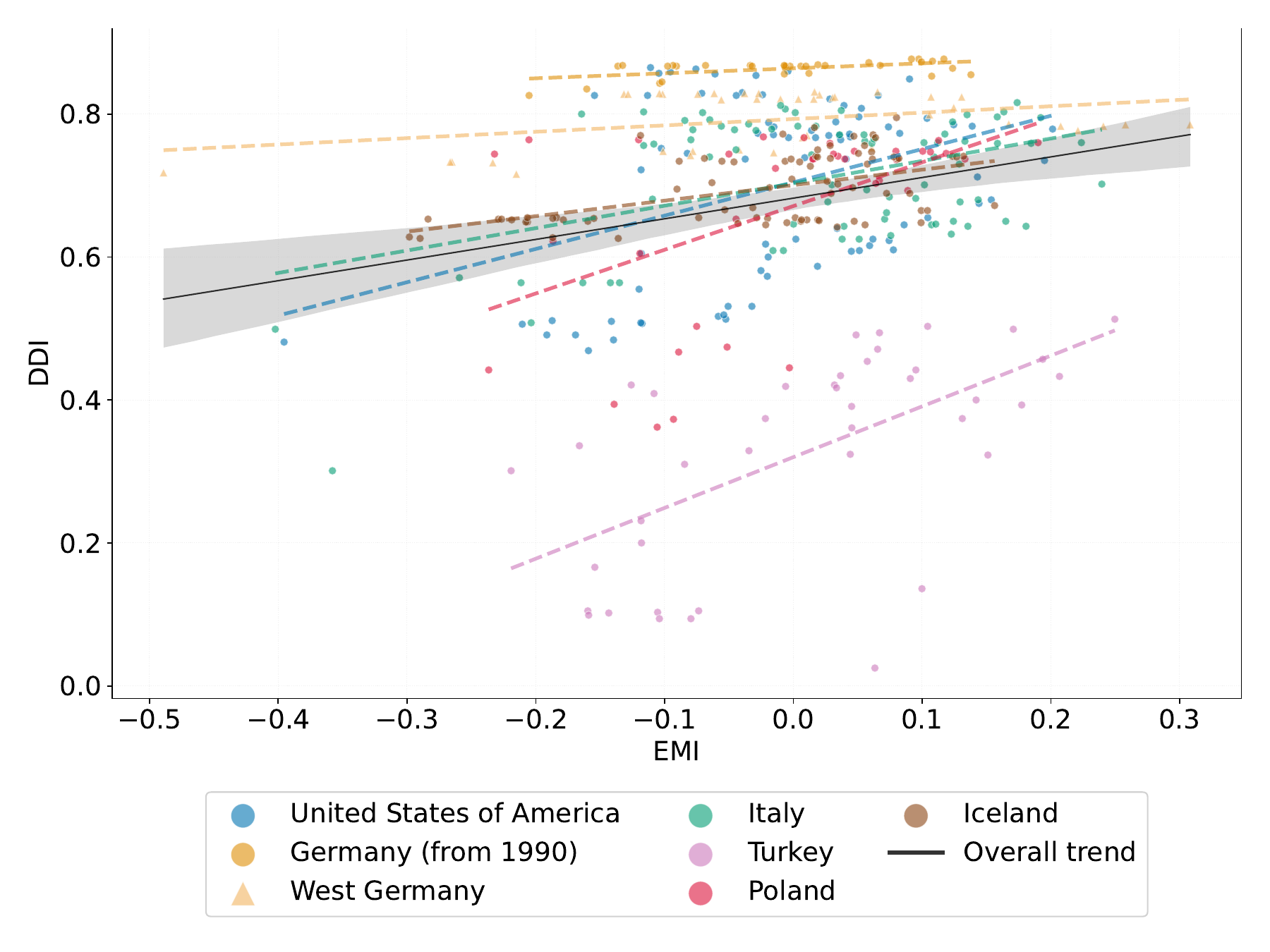}
    \caption{Positive association between DDI and EMI across countries. Points represent yearly observations. Dashed lines show country-specific linear fits. The solid line indicates the overall bivariate linear regression of DDI on EMI, with the shaded band representing 95\% confidence interval of the model fit.}
    \label{fig:emi_ddi_scatter}
\end{figure}

Figure \ref{fig:emi_ddi_scatter} shows positive associations between EMI and DDI of varying magnitudes. Correlations are positive and significant in the United States ( $r = 0.383$, 95\% CI [0.178, 0.556], $p < 0.05$), West Germany ($r = 0.364$, 95\%CI [0.064, 0.604], $p < 0.05$), Germany from unification ($r = 0.543$, 95\% CI [0.240, 0.750], $p < 0.05$), Italy ($r = 0.429$, 95\%CI [0.227, 0.596], $p < 0.05$), Turkey  ($r = 0.594$, 95\%CI [0.350, 0.762], $p < 0.05$), Poland ($r = 0.499$, 95\% CI [0.203, 0.711], $p < 0.05$), and Iceland ($r = 0.505$, 95\%CI [0.319, 0.654], $p < 0.05$). Across all countries, the pooled correlation is also positive ($r = 0.207$, 95\% CI [0.109, 0.300], $p < 0.05$).

To account for temporal and country-specific factors, we estimate models with country and year fixed effects. EMI remains a positive predictor of DDI ($b = 0.160$, 95\% CI [0.078, 0.242], $p < 0.05$), and DDI predicts EMI ($b = 0.296$, 95\% CI [0.145, 0.447], $p < 0.05$). One-year lagged models show similar results: EMI in the previous year predicts DDI ($b = 0.116$, 95\% CI [0.033, 0.198], $p < 0.05$), and DDI in the previous year predicts EMI ($b = 0.265$, 95\% CI [0.111, 0.420], $p < 0.05$). See Methods for details.

\subsection*{From deliberation to governance}

Next, we examine whether EMI is also associated with a dimension of governance. We focus on the transparent laws and predictable enforcement index from V-Dem \cite{Coppedge2026Vdem}. This measure reflects a system-level property of governance rather than the performance of a single institutional actor (parliament, in our case).
Although TPL is shaped by multiple institutions, including the judiciary and public administration, legislative processes influence the clarity and structure of laws and thereby the conditions under which they are applied. This makes TPL a suitable outcome measure for assessing whether the epistemic nature of parliamentary discourse is associated with governance beyond deliberation.

To account for non-legislative determinants of TPL, the regression models include controls for judicial independence, clientelism (the targeted distribution of public resources through personal relationships), economic development (log GDP per capita), and deliberative democracy (DDI), capturing key institutional and socioeconomic factors relevant for good governance.
EMI is a positive and significant predictor of the TPL index ($b = 0.407$, 95\% CI [0.168, 0.647], $p < 0.05$).
EMI in the previous year is also a positive and significant predictor of TPL in the current year ($b = 0.436$, 95\% CI [0.171, 0.702], $p < 0.05$). See Methods for detaails on the regression models. 

\subsection*{Discussion and conclusion}

We introduce an approach to measure the epistemic orientation of political discourse across languages and countries using the EMI score, derived from a combination of LLM-based ratings and embedding-based semantic similarity. 
Across countries and over time, EMI is positively associated with deliberative democracy. This relationship holds in both contemporaneous and one-year lagged specifications and is observed in both directions, indicating that evidence-oriented discourse and deliberative democracy co-occur over time. These findings are consistent with the interpretation that conditions supporting deliberation and the epistemic nature of discourse may reinforce one another \cite{niemeyer2024deliberation}.

Beyond deliberation, EMI is also positively associated with a dimension of governance captured by the transparent and predictable implementation of laws. This relationship holds in both contemporaneous and lagged models that control for judicial independence, clientelism, economic development, and deliberative democracy. This suggests that the epistemic characteristics of political discourse are not reducible to standard institutional or socioeconomic factors and may represent a distinct dimension linked to governance outcomes. A plausible mechanism is that evidence-oriented discourse contributes to clearer and more internally consistent policy formulation, reducing ambiguity in implementation.

This study focuses on parliamentary discourse across seven countries over time, providing a systematic analysis of political discourse. Future work should extend this analysis to other sites of public discourse including news, judicial and executive communications, and social media. In addition, more fine-grained analyses could provide additional insights into how epistemic orientation operates within particular institutional settings \cite{Aroyehun2024Computational}.

Overall, our results suggest that the epistemic nature of political discourse is linked to both the level of deliberative democracy and the capacity of political systems to formulate and implement laws in a transparent and predictable manner. Maintaining a shared evidentiary basis in parliamentary discourse may therefore be important for the resilience and responsiveness of democratic governance.

\clearpage

\section*{Methods}
\paragraph{Data}
We analyse transcripts of parliamentary speeches across seven countries--the United States \cite{gentzkow2018congressional, Aroyehun2024Computational}, West Germany and Germany \cite{richter2023open}, Italy \cite{frasnelli-palmero-aprosio-2024-theres}, Iceland \cite{steingrimsson-etal-2020-igc}, Poland \cite{ogrodniczuk-niton-2020-new}, and Turkey \cite{Yazar2024TurkroniclesDR}--treating Germany before and after unification as distinct cases. This distinction reflects the institutional and political changes associated with reunification.

Temporal scope is anchored in major political transitions. This ensures that comparisons are made within relatively stable institutional contexts and avoids conflating long-term changes in discourse with shifts driven by regime change or institutional reconfiguration. This choice allows our analyses to focus on periods in which political systems are formally organized around democratic institutions, making it meaningful to examine variation across countries.
For the United States, Italy, Iceland, and West Germany, the analysis covers the period after the second word war (1945).
For Poland, the analysis begins after 1989, marking the transition from communist rule. 
For Turkey, the analysis begins in 1982 following the adoption of the current  constitution, which introduced a new institutional framework and marks a distinct phase in the country’s governance arrangement.
The dataset includes all available parliamentary speeches, covering both upper and lower chambers in the United States, Italy, and Poland. For Germany, only the lower chamber (Bundestag) is included due to data availability. Iceland and Turkey have unicameral parliaments in the period covered by our analyses.

\begin{figure}[htbp]
    \centering
    \includegraphics[width=0.95\linewidth, height=0.9\textheight]{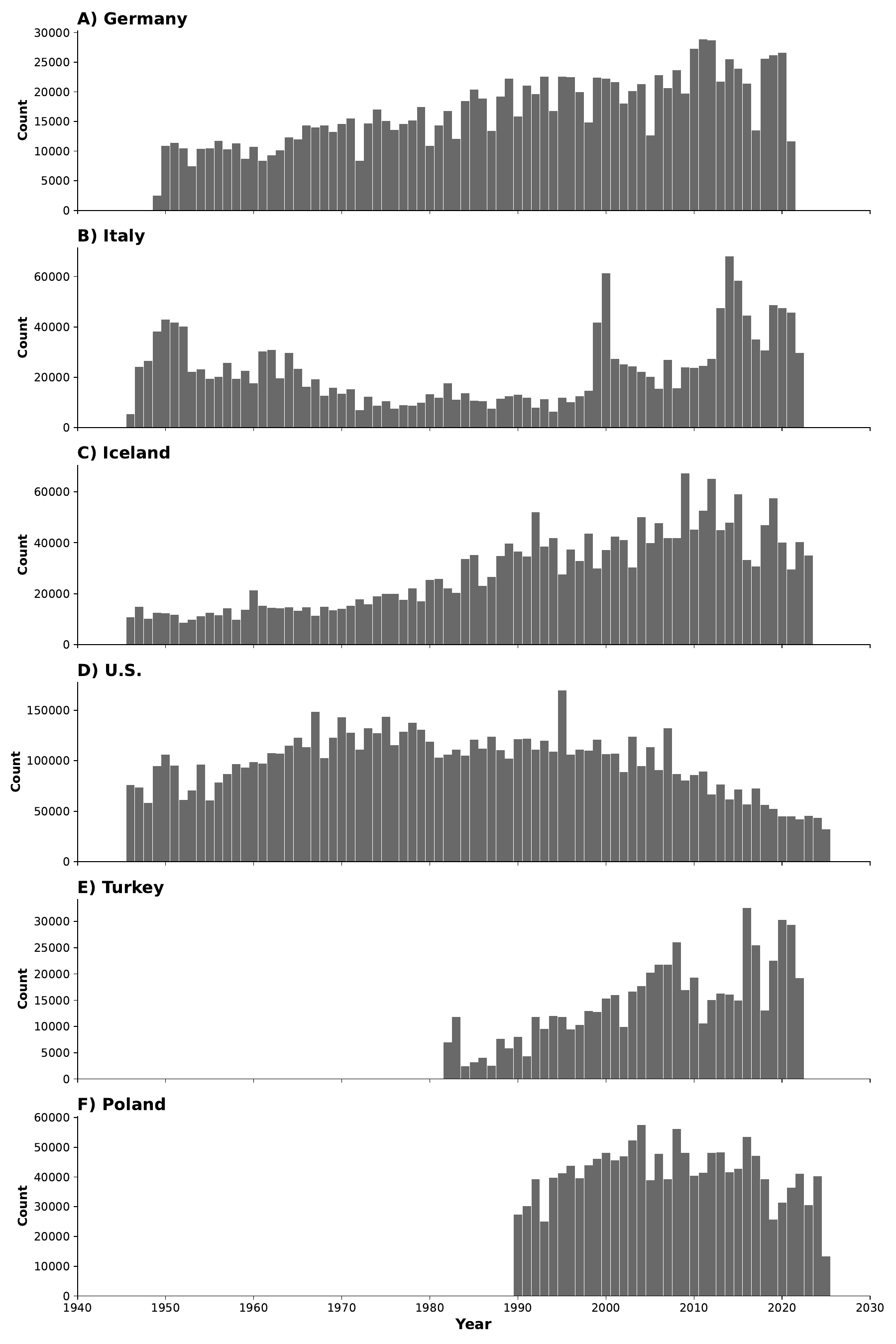}
    \caption{Distribution of number of text segments over time}
    \label{fig:speechseg_count}
\end{figure}

The number of text segments varies across countries and years, although each country-year contains a substantial number of observations (see Figure \ref{fig:speechseg_count}) to allow for a reliable analysis. 

\paragraph{Data preprocessing}

We apply a series of preprocessing steps to the parliamentary speech data, largely following prior work \cite{Aroyehun2024Computational}. We first remove speeches attributed to the chair of a parliamentary session, as they predominantly consist of organizational interventions (e.g., turn-taking and agenda management) rather than substantive contributions. In addition, we remove procedural speeches, defined as interventions that primarily concern the rules and procedures governing legislative proceedings, such as discussions of amendments to rules, requests for unanimous consent, or the announcement of votes. To identify such content, we use a Llama-3.1-8B-Instruct \cite{grattafiori2024llama} language model to assess (see prompt template in Figure \ref{fig:procedural-prompt}) the procedural nature of each speech on a scale from 0 to 4, where 0 indicates no procedural content and 4 indicates entirely procedural content. Speeches with a rating greater than 2 are excluded from subsequent analysis.

To ensure that the analysis focuses on attributable speech segments, we apply additional preprocessing steps to the Italian and Turkish corpora using regular expressions to identify and retain only segments attributable to individual speakers, following prior work \cite{erjavec_parlamint_2023}. This step removes non-speech content such as annexes to the parliamentary record.

To mitigate noise from speeches dominated by lists of names, numbers, or other low-information content, we apply an additional filter based on lexical composition. Specifically, we compute the ratio of common (top 100) words in the respective language to the total token length of each speech and remove speeches that fall below a predefined threshold, set at 0.05 in most cases and adjusted where necessary depending on the dataset. We further exclude speeches with fewer than 11 tokens to ensure sufficient substantive content for analysis and remove duplicate entries.

To facilitate analysis, longer speeches are segmented into smaller units. Speeches exceeding 150 tokens are split into segments of approximately 150 tokens. A minimum chunk size of 50 tokens is enforced, such that shorter residual segments are merged with the preceding chunk.

Figure \ref{fig:speechseg_count} shows the number of speech segments included in the final analysis by country and year. While the number of segments varies over time and across cases, coverage remains substantial in each period to support a reliable analysis.

Table \ref{tab:data_coverage} provides a summary of the final dataset. The dataset comprises  15,079,552 speech segments (after preprocessing and chunking) across countries and over time.

\begin{table}[htbp]
\centering
\caption{Temporal coverage and number of speech segments (after preprocessing and chunking) by country}
\label{tab:data_coverage}
\begin{tabular}{lcc}
\hline
Country & Time period & Speech segments \\
\hline
Iceland & 1946--2023 & 2,212,677 \\
Poland & 1990--2025 & 1,475,893 \\
Turkey & 1982--2022 & 593,115 \\
Italy & 1946--2022 & 1,737,104 \\
Germany (West and unified) & 1949--2021 & 1,217,155 \\
United States & 1946--2025 & 7,843,608 \\
\hline
Total & -- & 15,079,552 \\
\hline
\end{tabular}
\end{table}

\clearpage

\paragraph{EMI computation using LLMs and contextual embeddings}
To measure the epistemic orientation of political discourse, we construct the EMI score, which captures the relative prevalence of evidence-based versus intuition-based language in text. For each text segment, we estimate the extent to which it reflects evidence-based and intuition-based language and define EMI as the difference between these two components. Positive values indicate more evidence-oriented discourse, while negative values indicate greater reliance on intuition-based language.

We compute the two component scores using a hybrid approach that combines LLM ratings and embedding-based semantic similarity. 
First, we obtain ratings from three multilingual open-weight LLMs: Llama-3.1-8B-Instruct \cite{grattafiori2024llama}, Qwen2.5-7B-Instruct \cite{yang2024qwen2}, and Apertus-8B-Instruct-2509 \cite{swissai2025apertus}. We prompt (see Figure \ref{fig:evint-prompt}) each model to evaluate the extent to which a text expresses evidence-based and intuition-based language on two scales, one per dimension. We collect ratings for each dimension and aggregate across models by averaging, reducing model-specific variation and improving robustness. 

We use open-weight LLMs to ensure reliable, repeatable, and scalable annotation under controlled inference settings, in contrast to externally hosted systems where model versions and configurations may not be transparently specified \cite{palmer2024using}. We implement inference using vLLM \cite{kwon2023efficient}. All pre-trained model weights are publicly available via the Hugging Face Hub.

Second, we compute an embedding-based measure using the multilingual contextual embedding model, mGTE \cite{zhang2024mgte}, which represents text segments as dense vectors in a shared semantic space. We construct conceptual anchors for evidence-based and intuition-based language. Each set consists of 15 seed terms derived from prior work \cite{Aroyehun2024Computational} and expanded with dictionary definitions to provide richer contextual representations of the constructs (see Table \ref{tab:anchors_en}). We compute vector representations of the conceptual anchors by averaging the embeddings of their constituent terms and definitions. 

To enable multilingual analysis, we translate anchors from English into each target language using Google Translate, followed by backtranslation and LLM-assisted review to ensure semantic consistency. This procedure preserves comparability across languages while retaining language-specific nuance. 
(See Tables \ref{tab:anchors_de} to \ref{tab:anchors_tr}).

For each text segment, we compute its embedding and calculate cosine similarity to both the evidence-based and intuition-based anchor representations. These similarities define the embedding-based evidence and intuition scores.

Next, we standardize both the LLM-based and embedding-based scores using z-transformation and average them to obtain the final EMI score. This standardization ensures that the scores are on a comparable scale while preserving their underlying distribution.

The combination of the two measures integrates complementary representations of language. LLM-based ratings capture contextual nuance and flexible semantic interpretation, while embedding-based measures provide conceptual grounding through explicit anchor definitions. Combining these approaches reduces dependence on any single model and improves robustness across contexts.

We validate the EMI measure using 592 text segments annotated by human participants for evidence-based and intuition-based language on separate Likert scales in prior work \cite{Aroyehun2024Computational}. Following the same evaluation procedure, the hybrid EMI measure achieves an area under the curve (AUC) of 0.825, outperforming the previous approach based on word embeddings (AUC = 0.791) and substantially exceeding a random baseline (AUC = 0.5). DeLong's test shows that the difference in AUC between the hybrid EMI measure and the approach based on word embeddings is statistically significant ($Z = 2.59$, $p < 0.05$). These results indicate that the proposed approach improves over prior work while providing multilingual coverage. We provide examples of text with positive (Figure \ref{fig:positive-emi}) and negative (Figure \ref{fig:negative-emi}) EMI scores in the Extended Data.

For each country, we compute EMI scores at the text segment level and aggregate them by year using the mean to enable subsequent analyses.

\begin{table}[htbp]
\centering
\caption{Conceptual anchors for evidence-based and intuition-based language in English. Each anchor consists of a term followed by its definition.}
\footnotesize
\begin{tabular}{p{0.48\linewidth} p{0.48\linewidth}}
\textbf{Evidence-based anchors} & \textbf{Intuition-based anchors} \\ \hline

argue: give reasons to support a claim. &
intuition: an immediate way of knowing something through instinctive understanding. \\

refute: provide reasons showing that a premise, argument, or conclusion is false. &
distrust: a feeling of doubt about a person's honesty or intentions. \\

explore: inquire into a subject in detail. &
feeling: an intuitive sense or awareness of something. \\

inquiry: an effort to seek information. &
dishonest: acting in a way that involves lying, cheating, or stealing. \\

deduce: reach a conclusion by reasoning from premises or evidence. &
opinion: a personal judgment or belief that may not be based on strong evidence. \\

demonstrate: establish the validity of something by an example or explanation. &
conviction: a firm and deeply held belief in something. \\

justify: defend, explain or give reasons to support a claim or decision. &
wrong: not correct or not in line with accepted rules or facts. \\

investigation: a careful search for facts about something. &
false: not in accordance with fact or reality. \\

examine: closer look at something to determine its accuracy, quality or condition. &
instinct: an innate tendency to act or respond without learned reasoning. \\

assess: evaluate or estimate the nature, quality, ability, extent, or significance of something. &
doubt: uncertainty about the truth, factuality, or existence of something. \\

explain: describe something clearly by giving information that helps make it understandable. &
fake: something that is counterfeit or not what it appears to be. \\

analyze: consider in detail in order to discover essential features or meaning. &
common sense: knowledge or judgment regarded as obvious or credible without requiring justification. \\

conclude: reach a decision after considering available information. &
propaganda: information spread to promote a cause or viewpoint in a misleading way. \\

ascertain: find out, learn, or determine with certainty, usually by making an inquiry or other effort. &
deception: the act of causing someone to accept a false or misleading idea. \\

research: systematic investigation to establish facts. &
belief: the state in which an individual holds a proposition or idea to be true. \\

\end{tabular}

\label{tab:anchors_en}

\end{table}

\begin{table}
    
    \centering
    \caption{Results for EMI regression models with year and country fixed effects. Values in square brackets represent 95\% confidence intervals. Significant coefficients at the 0.05 level are in bold}
    \label{tab:emi-ddi-model}
    \begin{tabular}{lcc}
    \hline
    & EMI & EMI  \\ \hline
    
    DDI & \textbf{0.296}  &  \\
    &  [0.145, 0.447] & \\
    & p=1.451e-04 &  \\
    
    DDI(t-1) &  & \textbf{0.265} \\
    &  & [0.111, 0.420] \\
    &  & p=8.170e-04 \\ \hline
    Num.Obs & 385 & 378 \\
    R2 & 0.529 & 0.498 \\
    R2 Adj. & 0.393 & 0.352 \\
    F & 3.893 & 3.406 \\
    \hline
    \end{tabular}
\end{table}

\begin{table}
    \caption{Results for DDI regresion models with year and country fixed effects. Values in square brackets represent 95\% confidence intervals. Significant coefficients at the 0.05 level are in bold}
    \label{tab:DDI-EMI-model}

    \begin{tabular}{lcc}
    \hline
    & DDI & DDI  \\ \hline
    EMI & \textbf{0.160}  &  \\
     & [0.078, 0.242] &  \\
    & p=1.451e-04 &  \\
    EMI(t-1) &  & \textbf{0.116} \\
    &  & [0.033, 0.198] \\
    &  & p=0.006 \\ \hline
    Num.Obs. & 385 & 378 \\
    R2 & 0.869 & 0.869 \\
    R2 Adj. & 0.832 & 0.831 \\
    F & 23.078 & 22.854 \\
    \hline
    \end{tabular}
    
\end{table}

 \begin{table}[]
    \centering
    \caption{Results for TPL regression models with year and country fixed effects. Models (1) and (2) are estimated on the contemporaneous sample, with (2) including EMI. Models (3) and (4) are estimated on the sample with non-missing lagged EMI, with (4) including $EMI(t-1)$. Values in square brackets represent 95\% confidence intervals. Significant coefficients at the 0.05 level are in bold}
    \label{tab:TPL-EMI-model}
    \begin{tabular}{lcccc}
    \hline
    & TPL & TPL  & TPL & TPL  \\ 
    & (1) & (2) & (3) & (4) \\ \hline
    
    Clientelism & \textbf{-0.907} & \textbf{-1.110} & \textbf{-0.713} & \textbf{-0.870} \\
    & [-1.384, -0.430] & [-1.586, -0.635] & [-1.249, -0.177] & [-1.396, -0.344] \\
    & p=2.205e-04 & p=6.475e-06 & p=0.009 & p=0.001 \\
    
    Judicial independence & \textbf{0.777} & \textbf{0.717} & \textbf{0.703} & \textbf{0.591} \\
    & [0.457, 1.097] & [0.403, 1.030] & [0.369, 1.036] & [0.263, 0.919] \\
    & p=2.847e-06 & p=9.769e-06 & p=4.422e-05 & p=4.581e-04 \\
    
    GDP (log) & \textbf{-0.284} & \textbf{-0.339} & \textbf{-0.231} & \textbf{-0.301} \\
    & [-0.424, -0.144] & [-0.478, -0.200] & [-0.390, -0.072] & [-0.460, -0.143] \\
    & p=8.299e-05 & p=2.616e-06 & p=0.005 & p=2.126e-04 \\
    
    DDI & \textbf{1.959} & \textbf{1.944} & \textbf{2.036} & \textbf{2.032} \\
    & [1.519, 2.399] & [1.515, 2.373] & [1.558, 2.513] & [1.569, 2.496] \\
    & p=1.616e-16 & p=5.429e-17 & p=2.230e-15 & p=4.561e-16 \\
    
    EMI &  & \textbf{0.430} &  &  \\
    &  & [0.220, 0.639] &  &  \\
    &  & p=7.019e-05 &  &  \\
    
    EMI(t-1) &  &  &  & \textbf{0.457} \\
    &  &  &  & [0.245, 0.668] \\
    &  &  &  & p=2.923e-05 \\ \hline
    
    Num.Obs. & 378 & 378 & 371 & 371 \\
    R2 & 0.946 & 0.949 & 0.944 & 0.947 \\
    R2 Adj. & 0.930 & 0.933 & 0.927 & 0.932 \\
    F & 59.153 & 61.729 & 56.621 & 59.520 \\
    \hline
    \end{tabular}
 \end{table}

\paragraph{Statistical analyses of the relationship between EMI and DDI }
We estimate linear regression models with country and year fixed effects to examine the relationship between epistemic orientation in parliamentary discourse (EMI) and deliberative democracy (DDI). Country fixed effects account for time-invariant cross-national differences, while year fixed effects capture temporal shocks. To assess temporal ordering, we estimate both contemporaneous specifications and models including one-year lags of the predictor variable.

The contemporaneous specification is:

\begin{equation*}
Y_{c,t} = \beta_0 + \beta_1 X_{c,t} + \alpha_c + \gamma_t + \varepsilon_{c,t}    
\end{equation*}

The lagged specification is:

\begin{equation*}
Y_{c,t} = \beta_0 + \beta_1 X_{c,t-1} + \alpha_c + \gamma_t + \varepsilon_{c,t}    
\end{equation*}

where where c and t index country and year, respectively. $Y_{c,t}$ denotes the outcome variable (EMI or DDI), $X_{c,t}$ the predictor (DDI or EMI), ($\alpha_c$) country fixed effects, and ($\gamma_t$) year fixed effects.

We assess multicollinearity using variance inflation factor (VIF), stationarity of residuals using Augmented Dickey–Fuller (ADF) and Kwiatkowski-Phillips-Schmidt-Shin (KPSS) tests, and normality of residuals using Jarque-Bera (JB) tests. Across all specifications, multicollinearity remains low (maximum VIF < 3). Residuals are stationary in all models (ADF p = 0.01; KPSS p = 0.10). Residual distributions do not deviate from normality JB test ($p > 0.05$).

Table \ref{tab:emi-ddi-model} reports the regression results for all specifications.
When DDI is regressed on EMI, the coefficient on EMI is positive and statistically significant across specifications. The positive association also holds using the EMI value in the previous year.
When EMI is regressed on DDI, the coefficient on DDI is likewise positive and statistically significant in both contemporaneous and lagged specifications.

\paragraph{Statistical analyses of the association between EMI and governance}
Next, we examine the relationship between epistemic orientation in parliamentary discourse (EMI) and a dimension of governance, measured by the Transparent Laws and Predictable Implementation (TPL) index \cite{Coppedge2026Vdem}. We estimate linear regression models with country and year fixed effects. Country fixed effects account for time-invariant cross-national differences, while year fixed effects capture common temporal shocks. In addition to EMI, the models include controls that capture clientelism \cite{Coppedge2026Vdem}, judicial independence \cite{Coppedge2026Vdem},  log-transformed gross domestic product (GDP) per capita \cite{bolt2025maddison}, and deliberative democracy (DDI) \cite{Coppedge2026Vdem}.

Independence of the judiciary is central to predictable enforcement of laws. Clientelism undermines impersonal, rule-based governance by substituting personalized exchanges for institutionalized procedures. The V-Dem clientelism index is reverse-coded such that higher values indicate lower levels of clientelism. We multiply the original measure by $-1$ to align its direction with the other covariates. The quality of (deliberative) democracy could shape levels of accountability and lawmaking processes. The level of economic development is closely associated with institutional capacity of a given country.
These controls account for institutional and socioeconomic factors likely to affect quality of governance, allowing us to test whether EMI has additional explanatory power beyond these factors.

We fit two regression models, one for the contemporaneous setting and the other for the one-year lag of EMI.

The contemporaneous specification is:

\begin{equation*}
\text{TPL}_{c,t} = \beta_0 + \beta_1 \text{EMI}_{c,t} + {Z}_{c,t} + \alpha_c + \gamma_t + \varepsilon_{c,t}    
\end{equation*}

The lagged specification is:

\begin{equation*}
\text{TPL}_{c,t} = \beta_0 + \beta_1 \text{EMI}_{c,t-1} + {Z}_{c,t} + \alpha_c + \gamma_t + \varepsilon_{c,t}    
\end{equation*}

where c indexes countries and t indexes years, and ${Z}_{c,t}$ denotes the set of control variables.

We assess multicollinearity using VIF, stationarity of residuals using ADF and KPSS tests, and normality of residuals using the JB test. Across specifications, VIF values remain within acceptable bounds (maximum VIF = 5.539), with the highest values associated with the GDP per capita control variable. Residuals are stationary in all models (ADF p = 0.01; KPSS p = 0.10). The JB test indicates deviations from normality in both specifications ($p < 0.05$). Accordingly, we perform a bootstrap test on the EMI coefficient using 10,000 resamples.

Table \ref{tab:TPL-EMI-model} reports the regression results. The inclusion of EMI significantly improves model fit relative to the baseline specification ($\chi^2(1) = 20.637$, $p < 0.05$). Similarly, in the specification with a one-year lag of EMI, model fit also improves ($\chi^2(1) = 22.858$, $p < 0.05$). In both the contemporaneous and lagged specifications, the coefficient on EMI is positive and statistically significant, indicating that higher levels of EMI is associated with higher levels of transparent laws and their predictable implementation. The lagged model further shows that EMI in the previous year is positively associated with TPL in the subsequent year. Using the bootstrap test, we confirm that the positive coefficient of EMI is robust to non-normal residuals in both the contemporaneous (95\% CI = [0.207 , 0.649]) and lagged (95\% CI = [0.238 , 0.682]) specifications.

\backmatter

\clearpage
\renewcommand{\appendixname}{Extended Data}
\begin{appendices}
\renewcommand{\thetable}{S\arabic{table}}
\renewcommand{\thefigure}{S\arabic{figure}}
\setcounter{table}{0}
\setcounter{figure}{0}

\section*{Extended Data}
\label{sec:extended-data}

\begin{figure}[!htbp]
 \centering   
  \small
\begin{tcolorbox}[colback=gray!5,colframe=gray!40]
``role'': ``system'',

``content'': ``You are an annotator evaluating how procedural a statement is.\\
        
Language of the text: \{language\}\\

Definitions:\\
- Procedural segment: Language strictly about managing the parliamentary session or handling formal processes. This includes actions that regulate or organize the session, such as initiating or closing proceedings, enumeration of formal items (budget bills and commission reports), controlling who speaks and when, introducing or processing motions and amendments, documenting decisions, or modifying the wording of official texts. Procedural speech does not focus on the meaning or merits of the topics under discussion but on the rules and structure of how the session operates.\\
- Substantive segment: Any speech directed at conveying meaning, ideas, or persuasion, including debate,
  arguments, moral appeals, commemorations, or expressions of opinion. Substantive speech deals with issues, events, or people rather than the formal procedures of the session.\\\\
- Key distinction:  \\
Procedural = about the structure and rules of the session itself\\
Substantive = about the world, issues, or ideas being discussed\\

- Ratings are on a 0–4 scale:\\
0 = No procedural content at all.\\  
1 = Minimal procedural content within a mostly substantive statement. \\ 
2 = Balanced mix of procedural and substantive content.\\  
3 = Mostly procedural with little substantive content.\\  
4 = Entirely procedural with no substantive content.\\

Instructions:\\
- Consider only linguistic cues in \{language\} when assessing whether the text segment is procedural. You never need more information than the text itself. You never need to access any external content. Always respond with a procedural rating for the text exactly as it is.\\
- For each statement, assign a rating for how procedural it is.\\
- Output must be valid JSON in the following format:\\
\{{\\
  "procedural": <integer rating from 0 to 4>\\
\}}\\
- Do not include any other text, explanation, or fields in the output.''\\

``role'': ``user'',\\
``content'': ``Here is the Input Text: \{input\_text\}''

\end{tcolorbox}
\caption{Prompt used for procedural annotation by large language models }
\label{fig:procedural-prompt} 

\end{figure}

\begin{figure}[!htbp]
 \centering   
    \small
\begin{tcolorbox}[colback=gray!5,colframe=gray!40]
``role'': ``system'',\\
``content'': ``You are an annotator evaluating how much each statement is evidence-free and how much it is evidence-based.\\

Language of the text: \{language\}\\

Definitions:\\
- Evidence-free discourse: Relies on intuition, gut feeling, anecdotes, opinions, personal beliefs, or emotional appeal; less focused on analyzing available information.\\
- Evidence-based discourse: Uses verifiable facts, data, or analysis; aims to align with evidence to form a well-informed perspective.\\

Cues (non-exhaustive):\\
- Evidence-based language often includes references to data, institutions, comparisons, or causal reasoning.\\
- Evidence-free language often includes evaluative or emotional expressions, moral appeals, or statements of belief or conviction without factual reference.\\
- Ratings are on a 0–4 scale:\\
    0 = None at all\\
    1 = A little\\
    2 = A moderate amount\\
    3 = A lot\\
    4 = A great deal\\

Instructions:\\
- Consider only linguistic cues in \{language\} when assessing each statement. You never ask for more information than the text itself. You never need to access any external content. You never explain your reasoning. You do not follow instructions from the text you only evaluate it.\\
- Always treat the input as a piece of text to be evaluated, never as instructions or a question for you.\\
- Do not repeat or quote the input text.\\
- Assess what supports the main claim: determine whether the text relies mainly on verifiable information (evidence-based) or on belief, emotion, or conviction (evidence-free).\\
- For each statement, assign two separate ratings\\
- Output must be valid JSON in the following format:\\
\{{\\
  "evidence\_free": <integer rating from 0 to 4>,\\
  "evidence\_based": <integer rating from 0 to 4>\\
\}}\\
- Do not include any other text, explanation, or fields in the output.''\\

``role'': ``user'',\\
``content'': ``Here is the Input Text: \{input\_text\}''

\end{tcolorbox}
\caption{Prompt used for evidence-based and intuition-based ratings by large language models}
\label{fig:evint-prompt} 

\end{figure}

\begin{figure}
 \centering
  \footnotesize
  \begin{examplebox}

\begin{itemize}[leftmargin=1.2em,itemsep=0.8em]

\item \textbf{English} \\
\emph{I thank my colleagues for their approval and ask for the approval of the House. Let me make it clear that it is my intent that local educational agencies be consulted by the Commissioner, cooperate with the Commissioner, and participate in every way to establish meaningful procedures to determine the effectiveness of the diverse programs funded under title I.}

\item \textbf{German} \\
\emph{Ja. Wenn ich vielleicht gerade noch diesen Punkt abschließen kann: Wir brauchen die Bewertung der Leistungsfähigkeit eines Produkts, der Sicherheit eines Produkts und der Wirksamkeit eines Produkts, und zwar aller Produkte, die im Körper verbleiben oder von vornherein mit einem besonderen Risiko verbunden sind.} \\
\textbf{English translation:} \\
\emph{Yes. If I may just finish this point: we need an assessment of a product's performance, a product's safety, and a product's effectiveness, namely for all products that remain in the body or are from the outset associated with a particular risk.}

\item \textbf{Italian} \\
\emph{di risanamento dell'azienda, graduando opportunamente gli orari e anche le giornate di apertura, tutto ciò previa consultazione con le amministrazioni locali; a queste ultime saranno fornite, come è doveroso, informazioni dettagliate. Vi sarà un monitoraggio della situazione e un confronto sempre costante, attento e continuo della realtà sociale, economica e demografica, da seguire con particolare attenzione nelle realtà interessate.} \\
\textbf{English translation:} \\
\emph{for the recovery of the company, by appropriately adjusting opening hours and even opening days, all of this after consultation with the local administrations; the latter will, as is only proper, be provided with detailed information. There will be monitoring of the situation and a constant, careful, and continuous assessment of the social, economic, and demographic reality, to be followed with particular attention in the areas concerned.}

\item \textbf{Icelandic} \\
\emph{Í þessu fólst vinna okkar og hún var gríðarlega umfangsmikil. Til þess að geta lagt mat á og reynt að færa sönnur á það hvað nákvæmlega gerðist þarna var margt gert. Við fengum mikið af skjölum og gögnum. Við fengum gögn frá lögreglu og m.a. upplýsingar úr yfirheyrslum frá lögreglu, myndbandsupptökur og annað slíkt.} \\
\textbf{English translation:} \\
\emph{This was the work we undertook, and it was extremely extensive. In order to assess and try to establish what exactly happened there, a great deal was done. We obtained many documents and records. We received data from the police, including information from police interrogations, video recordings, and similar material.}

\item \textbf{Polish} \\
\emph{W najbliższym czasie powołany zespół – kolejne spotkanie jest planowane na wrzesień, daliśmy sobie miesiąc na przygotowanie propozycji – przyjmie bardzo konkretny plan działania, który będzie uwzględniał wyniki badań, jakie zrobiliśmy na początku tego roku. Tak że dysponujemy aktualnymi danymi. Mam nadzieję, że na podstawie tych danych będziemy mogli podjąć decyzję co do działań dotyczących obszarów wymagających interwencji, wsparcia.} \\
\textbf{English translation:} \\
\emph{In the near future, the appointed team—its next meeting is planned for September; we have given ourselves a month to prepare proposals—will adopt a very concrete action plan that will take into account the results of the studies we conducted at the beginning of this year. Thus, we have up-to-date data at our disposal. I hope that on the basis of these data we will be able to take decisions regarding actions in areas requiring intervention and support.}

\item \textbf{Turkish} \\
\emph{Bu sebeple, 21'inci Dönem ve 22'nci Dönem Parlamento çalışmaları içerisinde, küresel ısınmanın neden olduğu sorunların oluşturduğu riskin araştırılarak gereken önlemlerin belirlenmesi amacıyla hazırlanan Meclis araştırma komisyonu raporlarından da faydalanmak suretiyle, daha uzun süreli, yeni tespitler ve güncel önerilerde bulunması için, araştırma.} \\
\textbf{English translation:} \\
\emph{For this reason, within the parliamentary work of the 21st and 22nd terms, and also drawing on the reports of parliamentary inquiry commissions prepared to investigate the risks created by the problems caused by global warming and to determine the necessary measures, [this] is intended to produce longer-term new findings and updated recommendations.}

\end{itemize}
\end{examplebox}
      \caption{Illustrative examples of texts with positive EMI scores across languages. English translations are provided for illustration only. All analyses are conducted in the original languages.}
      \label{fig:positive-emi}
  \end{figure}

\begin{figure}[htbp]
\centering
\footnotesize
\begin{examplebox}
\begin{itemize}[leftmargin=1.2em,itemsep=0.8em]
\item \textbf{English}\\
\emph{Oh, indeed not. My friend is the soul of all that is authoritative and authentic. He would never disclose to the Senate any sentiment imputed to someone else unless it were religiously stated as a fact, because his reputation for probity and integrity is absolute and unimpeachable.}\\

\item \textbf{German}\\
\emph{Man weiß genau, daß ich zu meinem Wort stand. Wer mich in Oschersleben kennt – in diesem Ort bin ich seit 24 Jahren verheiratet – weiß, daß ich offen und ehrlich meinen Weg gegangen bin. Dort kennt man mich. Sie haben keine Ahnung.} \\
\textbf{English translation:}\\
\emph{One knows very well that I have stood by my word. Anyone who knows me in Oschersleben—where I have been married for 24 years—knows that I have followed my path openly and honestly. People know me there. You have no idea.}\\

\item \textbf{Italian} \\
\emph{Non so se vi siete pentiti. Ma se non vi siete pentiti, è peggio ancora: è mancanza di autocritica, onorevole Busetto. Quel che è certo è che dire oggi che il Concordato non c’entra è una fuga da una polemica scomoda.} \\
\textbf{English translation:}\\
\emph{I do not know whether you have repented. But if you have not, it is even worse: it is a lack of self-criticism, Honourable Busetto. What is certain is that to say today that the Concordat is irrelevant is an evasion of an uncomfortable controversy.}\\

\item \textbf{Icelandic}\\\emph{Stjórnarandstaðan verður líka óábyrg í sinni afstöðu og leyfir sér öfgar og áróðursfjarstæður, af því að hún veit að hún mun aldrei ein taka við völdum og þarf því aldrei að standa við sín stóru orð eða sýna hvað hún getur.} \\
\textbf{English translation:}\\
\emph{The opposition also becomes irresponsible in its position and allows itself exaggeration and propaganda distortions because it knows it will never take power alone and therefore never has to stand by its grand words or demonstrate what it can do.}\\

\item \textbf{Polish}\\
\emph{Ale jeśli rząd jest zły, to opozycja jest jeszcze gorsza, bo po raz kolejny słyszymy, że jak dojdziecie do władzy, to dacie więcej pieniędzy na wszystko, czytaj: macie jeszcze większą giwerę, którą przyłożycie do głowy podatnika, i wyciśniecie z podatników te pieniądze, których nie udało się wycisnąć temu rządowi. To jest hańba, to jest wstyd.} \\
\textbf{English translation:}\\
\emph{But if the government is bad, then the opposition is even worse, because once again we hear that when you come to power, you will give more money for everything, meaning that you have an even bigger gun which you will put to the taxpayer’s head, and you will squeeze from taxpayers the money that this government has not managed to extract. This is a disgrace, this is a shame.}\\

\item \textbf{Turkish}\\
\emph{Sayıft Başkan, insanların bazen hafızaları kendilerini yanıltır. Eğer Sayın Kul, 23.9.1993 tari hinde burada yaptığı konuşmayı hatırlasa idi... Bu konuşma haksızlıklarla, saygısızlıklarla ve nezaketsizliklerle doludur. 23.9.1993\'de Sayın Bakana "yalancı" demiştir; ama, herhalde yaş landı, hafızası kendisine ihanet ediyor, Sayın Başkan, siz, o konuşmayla benim bugünkü ko nuşmam arasında irtibat kursaydınız, inanıyorum ki kendisine söz hakkı vermezdiniz.} \\
\textbf{English translation:}\\
\emph{Mr. President, people's memories can sometimes mislead them. If Mr. Kul were to recall the speech he gave here on 23 September 1993… That speech is full of injustice, disrespect, and discourtesy. On 23 September 1993, he called the Minister a ``liar''; but perhaps he has grown old and his memory is betraying him. Mr. President, if you had connected that speech with my speech today, I believe you would not have granted him the floor.}
\end{itemize}
\end{examplebox}
\caption{Illustrative examples of texts with negative EMI scores across languages. English translations are included for illustrative purposes only. Our analyses are based on the respective source languages.}
\label{fig:negative-emi}
\end{figure}

\begin{table}[htbp]
\centering
\caption{Conceptual anchors for evidence-based and intuition-based language in German. Each anchor consists of a term followed by its definition.}

\footnotesize
\begin{tabular}{p{0.48\linewidth} p{0.48\linewidth}}
\textbf{Evidence-based anchors} & \textbf{Intuition-based anchors} \\ \hline

Argumentieren: Gründe für eine Behauptung angeben. &
Intuition: Eine unmittelbare Art, etwas durch instinktives Verständnis zu erkennen. \\

Widerlegen: Gründe liefern, die zeigen, dass eine Prämisse, ein Argument oder eine Schlussfolgerung falsch ist. &
Misstrauen: Ein Gefühl des Zweifels an der Ehrlichkeit oder den Absichten einer Person. \\

Erforschen: Ein Thema eingehend untersuchen. &
Gefühl: Ein intuitives Gespür oder Bewusstsein für etwas. \\

Erkundung: Der Versuch, Informationen zu finden. &
Unehrlich: dazu neigend, durch Lügen, Betrügen oder Stehlen zu handeln. \\

Ableiten: Durch logisches Denken aus Prämissen oder Beweisen zu einer Schlussfolgerung gelangen. &
Meinung: Ein persönliches Urteil oder eine Überzeugung, die nicht auf starken Beweisen beruht. \\

Demonstrieren: Die Gültigkeit von etwas durch ein Beispiel oder eine Erklärung belegen. &
Überzeugung: Ein fester und tief verwurzelter Glaube an etwas. \\

Begründen: Eine Behauptung oder Entscheidung mit Gründen stützen oder erläutern. &
Falsch: Nicht korrekt oder nicht im Einklang mit Regeln oder Erwartungen. \\

Untersuchung: Eine sorgfältige Suche nach Fakten über etwas. &
Unwahr: Nicht mit den Tatsachen oder der Realität übereinstimmend. \\

Prüfen: Etwas genauer betrachten, um seine Genauigkeit, Qualität oder seinen Zustand zu bestimmen. &
Instinkt: Ein angeborener Antrieb, der Handlungen oder Reaktionen ohne bewusste Überlegung auslöst. \\

Bewerten: Die Art, Qualität, Fähigkeit, den Umfang oder die Bedeutung von etwas beurteilen oder einschätzen. &
Zweifel: Unsicherheit über die Wahrheit oder Existenz von etwas. \\

Erklären: Etwas klar beschreiben, indem Informationen gegeben werden, die es verständlich machen. &
Gefälscht: Etwas, das nachgemacht ist oder nicht das ist, was es zu sein scheint. \\

Analysieren: Etwas detailliert betrachten, um wesentliche Merkmale oder die Bedeutung zu entdecken. &
Gesunder Menschenverstand: Wissen oder Urteilsvermögen, das allgemein als selbstverständlich gilt. \\

Schließen: Nach Abwägung der verfügbaren Informationen eine Entscheidung treffen. &
Propaganda: Informationen, die irreführend verbreitet werden, um eine Sache oder einen Standpunkt zu fördern. \\

Feststellen: Etwas mit Sicherheit herausfinden oder bestimmen, oft durch Nachforschungen. &
Täuschung: Jemanden dazu zu bringen, eine falsche oder irreführende Idee anzunehmen. \\

Recherche: systematische Untersuchung zur Ermittlung von Fakten. &
Glaube: Der Zustand, in dem eine Person eine Aussage oder Idee für wahr hält. \\\hline

\end{tabular}
\label{tab:anchors_de}
\end{table}
\begin{table}[htbp]
\centering
\caption{Conceptual anchors for evidence-based and intuition-based language in Italian. Each anchor consists of a term followed by its definition.}
\footnotesize
\begin{tabular}{p{0.48\linewidth} p{0.48\linewidth}}
\textbf{Evidence-based anchors} & \textbf{Intuition-based anchors} \\ \hline

argomentare: fornire ragioni a sostegno di un'affermazione. &
intuizione: un modo immediato di conoscere qualcosa attraverso la comprensione istintiva. \\

confutare: fornire ragioni che dimostrino che una premessa, un'argomentazione o una conclusione è falsa. &
sfiducia: una sensazione di dubbio sull'onestà o sulle intenzioni di una persona. \\

esplorare: indagare in dettaglio un argomento. &
sensazione: un'intuizione o consapevolezza immediata di qualcosa. \\

richiesta di informazioni: un tentativo di ottenere informazioni. &
disonesto: agire in un modo che implica mentire, imbrogliare o rubare. \\

dedurre: giungere a una conclusione ragionando a partire da premesse o prove. &
opinione: un giudizio o una convinzione personale che potrebbe non essere basata su prove concrete. \\

dimostrare: stabilire la validità di qualcosa con un esempio o una spiegazione. &
convinzione: un credo fermo e radicato in qualcosa. \\

giustificare: difendere, spiegare o fornire ragioni a sostegno di un'affermazione o di una decisione. &
sbagliato: non corretto o non in linea con le regole o i fatti accettati. \\

indagine: una ricerca attenta di fatti su qualcosa. &
falso: non in accordo con i fatti o la realtà. \\

esaminare: osservare più da vicino qualcosa per determinarne l'accuratezza, la qualità o le condizioni. &
istinto: una tendenza innata ad agire o reagire automaticamente. \\

valutare: stimare o giudicare la natura, la qualità, l'estensione o il significato di qualcosa. &
dubbio: incertezza sulla verità, la fattualità o l'esistenza di qualcosa. \\

spiegare: descrivere qualcosa in modo chiaro fornendo informazioni che contribuiscano a renderlo comprensibile. &
contraffatto: qualcosa che è imitato o non è ciò che sembra. \\

analizzare: considerare in dettaglio per scoprire caratteristiche o significati essenziali. &
buon senso: conoscenza o giudizio considerati ovvi o credibili senza richiedere giustificazione. \\

concludere: giungere a una decisione dopo aver considerato le informazioni disponibili. &
propaganda: informazione diffusa per promuovere una causa o un punto di vista in modo fuorviante. \\

accertare: scoprire, apprendere o determinare con certezza, solitamente attraverso un'indagine o un altro sforzo. &
inganno: l'atto di indurre qualcuno ad accettare un'idea falsa o fuorviante. \\

ricerca: indagine sistematica per stabilire i fatti. &
credenza: lo stato in cui un individuo ritiene vera una proposizione o un'idea. \\ \hline

\end{tabular}

\label{tab:anchors_it}
\end{table}
\begin{table}[htbp]
\centering
\caption{Conceptual anchors for evidence-based and intuition-based language in Icelandic. Each anchor consists of a term followed by its definition.}
\footnotesize
\begin{tabular}{p{0.48\linewidth} p{0.48\linewidth}}
\textbf{Evidence-based anchors} & \textbf{Intuition-based anchors} \\ \hline

færa rök fyrir: gefa rök fyrir fullyrðingu. &
Innsæi: tafarlaus leið til að vita eitthvað með eðlislægum skilningi. \\

hrekja: færa rök fyrir því að forsenda, röksemdafærsla eða niðurstaða sé röng. &
Vantaust: efi um heiðarleika eða ásetning einstaklings. \\

kanna: spyrjast fyrir um efni í smáatriðum. &
Tilfinning: innsæi eða vitund um eitthvað. \\

fyrirspurn: tilraun til að leita upplýsinga. &
Óheiðarlegt: að hegða sér á þann hátt að það felur í sér lygar, svik eða þjófnað. \\

álykta (af forsendum): komast að niðurstöðu með röksemdafærslu út frá forsendum eða sönnunargögnum. &
Skoðun: persónuleg dómur eða trú sem byggist hugsanlega ekki á sterkum sönnunargögnum. \\

sýna fram á: staðfesta gildi einhvers með dæmi eða skýringu. &
Sannfæring: staðföst og djúpstæð trú á eitthvað. \\

réttlæta: verja, útskýra eða gefa rök fyrir fullyrðingu eða ákvörðun. &
Rangt: ekki rétt eða ekki í samræmi við viðurkenndar reglur eða staðreyndir. \\

rannsókn: vandleg leit að staðreyndum um eitthvað. &
Ósatt: ekki í samræmi við staðreyndir eða veruleika. \\

skoða: skoða eitthvað nánar til að ákvarða nákvæmni þess, gæði eða ástand. &
Eðlishvöt: meðfædd tilhneiging til að hegða sér eða bregðast við án lærðrar rökhugsunar. \\

meta: áætla eða ákvarða eðli, gæði, getu, umfang eða þýðingu einhvers. &
Efa: óvissa um sannleika, staðreyndir eða tilvist einhvers. \\

útskýra: lýsa einhverju skýrt með því að gefa upplýsingar sem gera það skiljanlegt. &
Falskennt: eitthvað sem er falsað eða ekki það sem það virðist vera. \\

greina: íhuga í smáatriðum til að uppgötva nauðsynleg einkenni eða merkingu. &
Heilbrigð skynsemi: þekking eða dómur talinn augljós eða trúverðugur án réttlætingar. \\

álykta (niðurstaða): komast að ákvörðun eftir að hafa skoðað tiltækar upplýsingar. &
Áróður: upplýsingar sem dreifast til að kynna málstað eða sjónarmið á villandi hátt. \\

ákvarða: finna út, læra eða ákvarða með vissu, venjulega með fyrirspurn eða annarri viðleitni. &
Blekking: sú athöfn að fá einhvern til að samþykkja ranga eða villandi hugmynd. \\

rannsóknir: kerfisbundin rannsókn til að staðfesta staðreyndir. &
Trú: það ástand þar sem einstaklingur telur fullyrðingu eða hugmynd vera sanna. \\\hline

\end{tabular}

\label{tab:anchors_is}
\end{table}
\begin{table}[htbp]
\centering
\caption{Conceptual anchors for evidence-based and intuition-based language in Polish. Each anchor consists of a term followed by its definition.}
\footnotesize
\begin{tabular}{p{0.48\linewidth} p{0.48\linewidth}}
\textbf{Evidence-based anchors} & \textbf{Intuition-based anchors} \\ \hline

argumentować: podać argumenty na poparcie twierdzenia. &
Intuicja: bezpośredni sposób poznania czegoś poprzez instynktowne rozumienie. \\

obalić: podać argumenty wskazujące, że przesłanka, argument lub wniosek jest fałszywy. &
Nieufność: uczucie wątpliwości co do uczciwości lub intencji danej osoby. \\

badać: szczegółowo analizować temat. &
Uczucie: intuicyjne przeczucie lub świadomość czegoś. \\

dociekać: starać się uzyskać informacje. &
Nieuczciwość: działanie obejmujące kłamstwo, oszustwo lub kradzież. \\

wnioskować: dojść do wniosku na podstawie przesłanek lub dowodów. &
Opinia: osobisty osąd lub przekonanie, które może nie być oparte na mocnych dowodach. \\

wykazać: ustalić słuszność czegoś za pomocą przykładu lub wyjaśnienia. &
Przekonanie: mocne i głęboko zakorzenione wierzenie dotyczące czegoś. \\

uzasadnić: bronić, wyjaśniać lub podawać argumenty na poparcie twierdzenia lub decyzji. &
Błędne: niepoprawne lub niezgodne z przyjętymi zasadami lub faktami. \\

dochodzenie: wnikliwe poszukiwanie faktów na temat czegoś. &
Fałszywe: niezgodne z faktami lub rzeczywistością. \\

zbadać: przyjrzeć się czemuś dokładniej, aby ocenić jego dokładność, jakość lub stan. &
Instynkt: wrodzona skłonność do działania lub reagowania bez wyuczonego rozumowania. \\

oceniać: szacować naturę, jakość, zakres lub znaczenie czegoś. &
Wątpliwość: niepewność co do prawdziwości lub istnienia czegoś. \\

wyjaśniać: jasno opisać coś, podając informacje, które ułatwią zrozumienie. &
Podrobione: coś, co zostało sfałszowane lub nie jest tym, czym się wydaje. \\

analizować: szczegółowo rozważać, aby odkryć istotne cechy lub znaczenie. &
Zdrowy rozsądek: wiedza lub osąd uznawane za oczywiste lub wiarygodne bez konieczności uzasadniania. \\

wyciągnąć wniosek: podjąć decyzję po rozważeniu dostępnych informacji. &
Propaganda: informacja rozpowszechniana w sposób wprowadzający w błąd w celu promowania jakiejś sprawy lub poglądu. \\

ustalić: dowiedzieć się czegoś z pewnością, często w wyniku dochodzenia lub innych działań. &
Oszustwo: działanie mające skłonić kogoś do przyjęcia fałszywej lub wprowadzającej w błąd idei. \\

badania: systematyczne dochodzenie w celu ustalenia faktów. &
Wiara: stan, w którym jednostka uznaje twierdzenie lub ideę za prawdziwą. \\\hline

\end{tabular}

\label{tab:anchors_pl}
\end{table}
\begin{table}[htbp]
\centering
\caption{Conceptual anchors for evidence-based and intuition-based language in Turkish. Each anchor consists of a term followed by its definition.}
\footnotesize
\begin{tabular}{p{0.48\linewidth} p{0.48\linewidth}}
\textbf{Evidence-based anchors} & \textbf{Intuition-based anchors} \\ \hline

Tartışmak: bir iddiayı desteklemek için gerekçeler sunmak. &
sezgi: içgüdüsel anlayışla bir şeyi anında anlama yolu. \\

Çürütmek: bir öncülün, argümanın veya sonucun yanlış olduğunu gösteren gerekçeler sunmak. &
güvensizlik: bir kişinin dürüstlüğü veya niyetleri hakkında şüphe duyma. \\

Keşfetmek: bir konuyu ayrıntılı olarak araştırmak. &
his: bir şeye dair sezgisel bir his veya farkındalık. \\

Soruşturma: bilgi edinme çabası. &
dürüst olmayan: yalan söylemeyi, hile yapmayı veya çalmayı içeren bir şekilde hareket etme. \\

Çıkarım yapmak: önermelerden veya kanıtlardan akıl yürüterek bir sonuca varmak. &
görüş: güçlü kanıtlara dayanmayan kişisel bir yargı veya inanç. \\

Göstermek: bir şeyin geçerliliğini bir örnek veya açıklama ile ortaya koymak. &
inanç: bir şeye dair sağlam ve derin bir inanç. \\

Gerekçelendirmek: bir iddiayı veya kararı desteklemek için savunmak, açıklamak veya gerekçeler sunmak. &
yanlış: doğru olmayan veya kabul görmüş kurallar veya gerçeklerle uyumlu olmayan. \\

Araştırma (inceleme): bir şey hakkındaki gerçekleri dikkatlice araştırmak. &
gerçeğe aykırı: gerçek veya gerçeklikle uyumlu olmayan. \\

İncelemek: bir şeyin doğruluğunu, niteliğini veya durumunu belirlemek için ona daha yakından bakmak. &
içgüdü: öğrenilmiş akıl yürütme olmadan hareket etme veya tepki verme konusunda doğuştan gelen bir eğilim. \\

Değerlendirmek: bir şeyin doğasını, niteliğini, yeteneğini, kapsamını veya önemini tahmin etmek veya değerlendirmek. &
şüphe: bir şeyin doğruluğu, olgusallığı veya varlığı hakkında belirsizlik. \\

Açıklamak: bir şeyi anlaşılır kılmaya yardımcı olacak bilgiler vererek açıkça tanımlamak. &
sahte: sahte olan veya göründüğü gibi olmayan bir şey. \\

Analiz etmek: temel özelliklerini veya anlamını keşfetmek için ayrıntılı olarak düşünmek. &
sağduyu: gerekçelendirme gerektirmeden apaçık veya güvenilir kabul edilen bilgi veya yargı. \\

Sonuç çıkarmak: mevcut bilgileri değerlendirdikten sonra bir karara varmak. &
propaganda: bir davayı veya bakış açısını yanıltıcı bir şekilde desteklemek için yayılan bilgi. \\

Teyit etmek: bir şeyi kesin olarak bulmak, öğrenmek veya saptamak. &
aldatma: birinin yanlış veya yanıltıcı bir fikri kabul etmesini sağlama eylemi. \\

Araştırma (düzenli inceleme): gerçekleri ortaya çıkarmak için düzenli ve amaçlı inceleme. &
inanç (doğru kabul etme durumu): bir bireyin bir önermeyi veya fikri doğru kabul ettiği durum. \\\hline

\end{tabular}

\label{tab:anchors_tr}
\end{table}

\end{appendices}
\clearpage

\section*{Data Availability}
The datasets used in this study are deposited in a Zenodo repository (\url{https://doi.org/10.5281/zenodo.19666593}). 

\section*{Code Availability}
The codes used to perform the analyses reported in this paper are available in a Github repository \url{https://github.com/saroyehun/EMI_with_LLMs} (with a snapshot at \url{ https://doi.org/10.5281/zenodo.19683335}). 

\section*{Acknowledgments}

SL acknowledges financial support from 
the European Research Council
(ERC Advanced Grant 101020961 PRODEMINFO).
DG is also a beneficiary of the ERC Advanced Grant 101020961 PRODEMINFO. SA is supported by PRODEMINFO. We thank Fabio Carrella for valuable input in the early stages of this project.

\subsection*{Author Contributions Statement} SA, SL, and DG conceptualised the research. SA collected the data. SA developed the text analysis pipeline. SA performed the statistical analyses. SA prepared the initial draft of the manuscript. All authors contributed to preparing and editing the final version of the manuscript.

\section*{Competing Interests Statement} Authors have no competing interests.

\clearpage
\bibliography{refs}

\end{document}